\documentclass[10pt,twocolumn,letterpaper]{article}
\def\confName{CVPR}
\usepackage[pagenumbers]{cvpr}            % To produce the CAMERA-READY version
\usepackage{multirow}
\usepackage{booktabs}
\usepackage{amsmath,amssymb,amsfonts}
\usepackage{graphicx}
\usepackage{textcomp}
\usepackage{tabularx}
\usepackage{xcolor}
\usepackage{algorithm}
\usepackage{stfloats}
\usepackage{algpseudocode}
\usepackage{svg}
\usepackage{pgfplots}
\pgfplotsset{compat=1.18}
\usepgfplotslibrary{groupplots}
\usepackage{makecell}
\usepackage{tikz}

\definecolor{cvprblue}{rgb}{0.21,0.49,0.74}
\usepackage[pagebackref,breaklinks,colorlinks,allcolors=cvprblue]{hyperref}

\begin{document}
\title{Joint Multi-Camera LiDAR Extrinsic Calibration via Learned Pairwise Initialization and Geometric Refinement}

\author{
    Aziz Al-Najjar, Marzieh Amini, James R. Green, and Felix Kwamena\\
   Department of System and Computer Engineering, Carleton University, Ottawa, Canada% % ional Sciences and Engineering Research Council (NSERC) of Canada and by the Infrastructure Resilience Research Group (IRRG).}
}

\maketitle

\begin{abstract}
Most learning-based camera--LiDAR calibration methods treat each camera--LiDAR pair independently, ignoring the rigid geometric coupling in multi-camera platforms. As a result, per-camera estimates may be individually accurate yet inconsistent at the system level. We present a two-stage framework for joint multi-camera LiDAR extrinsic calibration that combines learned pairwise matching with geometric refinement. First, CMRNext is applied independently to each camera to produce initial extrinsic estimates and dense 2D--3D correspondences. These predictions are then jointly refined through a multi-frame bundle adjustment with reprojection, per-camera prior, and relative-pose prior terms. This approach converts pairwise predictions into a globally consistent multi-camera calibration. Experiments on KITTI (in-domain for CMRNext) and Walkley (out-of-domain) datasets show improved per-camera accuracy and inter-camera consistency. On KITTI, the method achieves 0.89\,cm translation error and 0.038$^\circ$ rotation error. On Walkley, it reduces translation error from 108.6\,cm to 3.1\,cm, highlighting the benefit of explicit multi-camera coupling when single-camera predictions are less reliable.
\end{abstract}

\noindent\textbf{Keywords:} camera-LiDAR calibration, multi-camera calibration, extrinsic calibration, sensor fusion, bundle adjustment, geometric refinement, cross-modal matching.    

\section{Introduction}

Modern perception systems rely on combining the geometric accuracy of LiDAR with the rich visual information of cameras to support tasks such as object detection, semantic segmentation, and scene understanding~\cite{al2024identifying, cheng2025improving}. This fusion critically depends on accurate extrinsic calibration, which defines the mapping that projects 3D LiDAR points into the camera image plane. Even small calibration errors can lead to significant reprojection misalignment, degrading downstream performance \cite{pun2025plant, al2025lineshield}.

While single camera-LiDAR calibration has been studied extensively \cite{zhou2018lineplane,verma2019pointplane,pandey2012targetless,pandey2015mutual,schneider2017regnet,iyer2018calibnet,lv2021lccnet,cattaneo2025cmrnext}, many real-world platforms deploy multiple cameras around a single LiDAR sensor to increase field of view and robustness \cite{geiger2013kitti, OurData}. In such systems, the calibration problem acquires an additional geometric structure: all cameras are rigidly attached to the same platform, and their extrinsics are therefore not independent. Instead, they are coupled through the shared LiDAR coordinate frame, which fully determines the relative pose between cameras. Ignoring this coupling can result in calibrations that are individually plausible but jointly inconsistent, leading to misalignment artifacts across views \cite{huang2025whatreallymatters}.

This raises the central question addressed in this work: \emph{can learned pairwise camera--LiDAR calibration be improved by jointly optimizing all cameras under the rigid geometric constraints of a shared multi-camera platform?}

To answer this question, we propose a two-stage framework for joint multi-camera LiDAR extrinsic calibration. In Stage~1, a single-camera learning-based method, CMRNext, is applied independently to each camera to produce initial extrinsic estimates alongside dense 2D--3D correspondences. In Stage~2, these outputs are jointly refined through a multi-frame bundle adjustment that combines a reprojection term with per-camera and relative-pose priors to enforce inter-camera consistency.

We evaluate the proposed method on two datasets with complementary characteristics. On KITTI~\cite{geiger2013kitti}, one of the datasets on which CMRNext was originally trained and performed well. On Walkley~\cite{OurData}, a custom outdoor multi-camera LiDAR dataset with a different sensor configuration and image resolution, the CMRNext predictions are substantially less reliable, and the benefits of joint refinement become much larger. This contrast allows us to study whether multi-camera coupling helps, and under which operating conditions it provides improvements.

The main contributions of this work are:
\begin{itemize}
    \item A joint formulation for multi-camera LiDAR extrinsic calibration that uses learned single-camera predictions as initialization and refines all camera extrinsics together, rather than independently.
    \item An explicit relative-pose prior that enforces inter-camera geometric consistency during optimization and enables information to propagate across cameras.
    \item A study on in-domain and out-of-domain datasets showing when reprojection-only refinement is sufficient and when adding per-camera and relative-pose priors yields additional gains in both per-camera accuracy and system-level consistency.
\end{itemize}

\section{Related Work}

Camera-LiDAR calibration methods can be broadly divided into target-based, targetless, and learning-based approaches.

\paragraph{Target-based calibration:}
Target-based methods use known geometric structures such as checkerboards or planar targets to establish explicit 2D-3D correspondences. For example, Zhou \emph{et al.}~\cite{zhou2018lineplane} exploit line and plane constraints extracted from a calibration board, while Verma \emph{et al.}~\cite{verma2019pointplane} combine point and plane correspondences to solve for the rigid transform. These methods typically reduce calibration to a Perspective-$n$-Point (PnP) or least-squares problem with well-conditioned constraints, yielding high accuracy. However, they require controlled data collection and manual setup, limiting their practicality in dynamic or large-scale deployments.

\paragraph{Targetless calibration:}
Targetless methods avoid calibration objects by leveraging natural scene structure. A classical family of approaches formulates calibration as an information-theoretic alignment problem, maximizing mutual information between LiDAR-derived attributes (e.g., depth or reflectance) and image intensities~\cite{pandey2012targetless,pandey2015mutual}. Other methods rely on geometric consistency, such as aligning planes or edges across modalities, or minimizing reprojection error through bundle-adjustment-style optimization~\cite{borer2024chaos,li2023planeba}. While these methods eliminate the need for calibration targets, they often depend on scene geometry, good initialization, or handcrafted features, and can struggle in sparse or ambiguous environments.

\paragraph{Learning-based calibration:}
Learning-based approaches aim to improve robustness through data-driven feature extraction. Regression-based methods such as RegNet, CalibNet, and LCCNet directly predict the extrinsic transformation from RGB images and projected LiDAR inputs~\cite{schneider2017regnet,iyer2018calibnet,lv2021lccnet}. These models typically use convolutional encoders and optimize geometric consistency losses, but they implicitly learn dataset-specific cues and may generalize poorly across sensor configurations~\cite{huang2025whatreallymatters}.

More recent matching-based methods decouple correspondence estimation from geometric optimization. CMRNext~\cite{cattaneo2025cmrnext} follows this paradigm by predicting dense optical flow between a rendered LiDAR depth map and the camera image, from which 2D-3D correspondences are derived and used to estimate the extrinsic. This formulation improves generalization, but it remains limited to single camera-LiDAR pairs and does not enforce consistency across multiple cameras.

\paragraph{Multi-sensor calibration:}

Multi-sensor calibration methods extend the problem beyond a single camera-LiDAR pair and aim to recover a globally consistent set of extrinsics. Tu \emph{et al.}~\cite{tu2022multicamera} formulate multi-camera-LiDAR calibration as a joint structure-from-motion problem. UniCal~\cite{yang2024unical} learns calibration in a unified framework across heterogeneous sensors, while continuous-time formulations jointly model calibration and sensor motion over trajectories~\cite{lv2025continuous}. Multi-Calib~\cite{hu2025multicalib} further studies scalable calibration under varying sensor configurations. These methods address global consistency, but they do not exploit the recent strength of learned pairwise camera-LiDAR matching as the front end of a rig-aware refinement stage.

\section{Methodology}

\begin{figure*}
    \centering
    \includegraphics[width=0.85\linewidth]{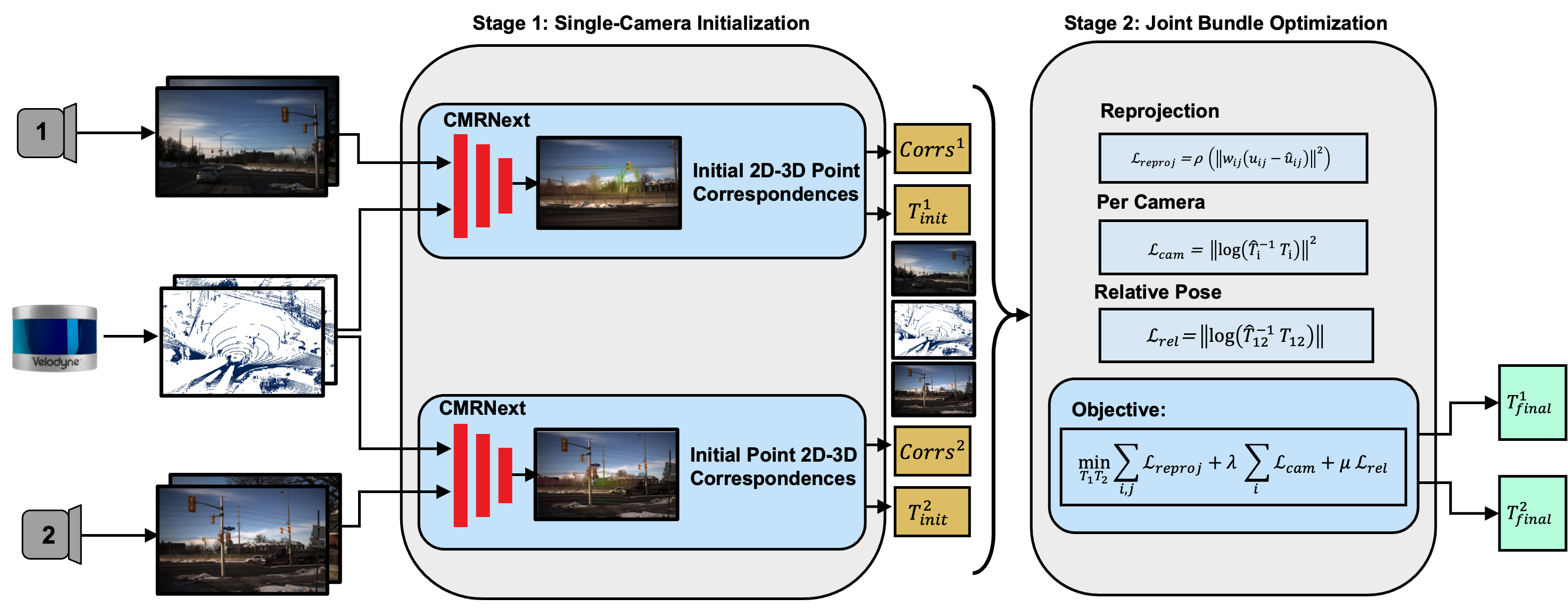}
    \caption{Overview of the proposed joint multi-camera LiDAR calibration framework}
    \label{fig:method}
\end{figure*}

Our proposed pipeline first obtains per-camera extrinsic initialization from CMRNext and then jointly refines all extrinsics using multi-frame BA (Fig.~\ref{fig:method}).

\subsection{Problem Formulation}

Let $\mathcal{F}_L$ denote the LiDAR coordinate frame, and let 
$\{\mathcal{F}_{C_1}, \ldots, \mathcal{F}_{C_N}\}$ denote the coordinate frames of 
$N$ rigidly mounted cameras, where $C_i$ is the $i$-th camera. The goal is to estimate the set of extrinsic transformations
\begin{equation}
    \mathcal{T} = \bigl\{T_{L\to C_1},\, \ldots,\, T_{L\to C_N}\bigr\},
    \quad T_{L\to C_i} \in SE(3)
\end{equation}
where $SE(3)$ denotes the special Euclidean group of 3D rigid-body transformations, and each
$T_{L\to C_i}$ maps a 3D point from the LiDAR frame to the coordinate frame of camera $C_i$.

Given a synchronized LiDAR point cloud $\mathbf{P}^t = \{\mathbf{p}_k^t\}$ and camera images $\{I_i^t\}$ at time $t$, a 3D point $\mathbf{p}_k^t$ projects into camera $C_i$ through the corresponding camera extrinsic and intrinsic model:
\begin{equation}
    \hat{\mathbf{u}}_{ik}^t = \pi_i\!\left(T_{L\to C_i}\,\mathbf{p}_k^t\right)
\end{equation}
where $\pi_i$ denotes perspective projection with intrinsic matrix $K_i$. Calibration is accurate when the projected point $\hat{\mathbf{u}}_{ik}^t$ aligns with its observed 2D correspondence $\bar{\mathbf{u}}_{ik}^t$ across all cameras and frames.

The distinguishing feature of the multi-camera setting is that the extrinsics are coupled by the shared rigid-body structure of the multi-camera rig. In particular, the relative pose between any two cameras $C_i$ and $C_j$ is not arbitrary; it is fully determined by their LiDAR-to-camera extrinsics:
\begin{equation}
    T_{C_i \to C_j} = T_{L\to C_j}\;T_{L\to C_i}^{-1}.
\end{equation}

This coupling motivates a joint refinement over all camera extrinsics.

\subsection{Stage 1: Learned Single-Camera Initialization}

CMRNext method~\cite{cattaneo2025cmrnext} has demonstrated solid results on KITTI and in-house datasets. Hence, we selected it as the single-camera initialization module. For each synchronized image-point-cloud pair $(I_i^t, \mathbf{P}^t)$, CMRNext predicts dense optical flow between a rendered LiDAR depth image and the camera image. From this cross-modal matching process, it extracts a set of 2D-3D correspondences $\{(\bar{\mathbf{u}}_{ik}, \mathbf{p}_k)\}$ and estimates a per-frame camera-LiDAR extrinsic $\hat{T}_{L\to C_i}^t$.

For each camera, Stage~1 provides an initial extrinsic estimate and confidence-weighted 2D--3D correspondences for Stage~2.

To obtain a single robust initialization for each camera, we aggregate the per-frame predictions across the sequence by taking the median, yielding
\begin{equation}
    \hat{T}_{L\to C_i}.
\end{equation}

\subsection{Stage 2: Joint Multi-Camera BA}

Building on the CMRNext predictions for each camera-LiDAR pair, this stage refines the calibrations further by fusing the information from other cameras in the system.

\subsubsection{Parameterization}

Each extrinsic $T_i \in SE(3)$ is parameterized by a 6-vector
\begin{equation}
    x_i = [\boldsymbol{\omega}_i^\top,\, \mathbf{t}_i^\top]^\top \in \mathbb{R}^6
\end{equation}
where $\boldsymbol{\omega}_i \in \mathbb{R}^3$ is an axis-angle rotation vector and $\mathbf{t}_i \in \mathbb{R}^3$ is the translation vector. The corresponding rigid transformation is recovered via the $SE(3)$ exponential map:
\begin{equation}
    T_i = \exp(x_i)
\end{equation}

For the two-camera case considered here, the joint optimization state is
\begin{equation}
    x = \begin{bmatrix} x_1^\top & x_2^\top \end{bmatrix}^\top \in \mathbb{R}^{12}
\end{equation}
initialized from the Stage-1 extrinsic estimates as
\begin{equation}
    x_0 = [\log(\hat{T}_1)^\top \mid \log(\hat{T}_2)^\top]^\top
\end{equation}

\subsubsection{Optimization Objective}

The joint refinement is formulated as a three-term objective over all accumulated frames and correspondences:
\begin{equation}
\label{eq:objective}
\begin{aligned}
\min_{T_1, T_2}\;&
\underbrace{\sum_{i=1}^{2}\sum_{k} \rho\!\left(\bigl\|\mathbf{w}_{ik}(\bar{\mathbf{u}}_{ik} - \pi_i(T_i,\,\mathbf{p}_k))\bigr\|^2\right)}_{\text{reprojection}} \\
&+\;\underbrace{\lambda\sum_{i=1}^{2}\bigl\|\log\!\left(\hat{T}_i^{-1} T_i\right)\bigr\|^2}_{\text{per-camera prior}}
+\;\underbrace{\mu\,\bigl\|\log\!\left(\hat{T}_{12}^{-1}\,T_2 T_1^{-1}\right)\bigr\|^2}_{\text{relative-pose prior}} 
\end{aligned}
\end{equation}
Here,
\begin{equation}
    \hat{T}_{12} = \hat{T}_2\,\hat{T}_1^{-1}
\end{equation}
is the inter-camera relative pose implied by the Stage-1 estimates, and $\rho(\cdot)$ is a robust loss function.
Each term in Eq.~\eqref{eq:objective} serves a distinct role.
The first term minimizes reprojection error over all correspondences, encouraging each LiDAR point to align with its observed image location under the current extrinsics. The per-camera prior stabilizes the optimization when correspondences are sparse or noisy by keeping each extrinsic near its Stage-1 estimate $\hat{T}_i$. The relative-pose prior ties the two camera extrinsics together by favoring consistency with the Stage-1 inter-camera geometry $\hat{T}_{12}$. Together, these terms combine data fitting with per-camera and inter-camera geometric regularization.

\subsubsection{Robust Loss}

The correspondence set produced by Stage~1 may still contain residual outliers or inaccurate matches. To reduce the influence of such errors during refinement, we apply a Cauchy robust loss to the squared reprojection residual:
\begin{equation}
    \rho(s) = \delta^2 \log\!\left(1 + \frac{s}{\delta^2}\right)
\end{equation}
where
\begin{equation}
    s = \bigl\|\bar{\mathbf{u}}_{ik} - \pi_i(T_i,\,\mathbf{p}_k)\bigr\|^2 
\end{equation}
Here, \(\delta\) is the robust-loss scale, expressed in pixels. The Cauchy loss down-weights large residuals more aggressively, which is beneficial in our setting as the learned correspondence predictions may include a non-negligible fraction of large-error matches.

\subsubsection{Correspondence Preprocessing}

Since the success of joint refinement depends on the quality and spatial distribution of the Stage-1 correspondences, we perform several preprocessing steps before optimization:
\begin{enumerate}
    \item \textbf{Depth filter}: correspondences whose 3D points satisfy $Z \leq 0$ under the current extrinsic are discarded.
    \item \textbf{Confidence pre-filter}: correspondences with confidence below a minimum threshold $c_\text{min}$ are removed.
    \item \textbf{Spatial subsampling}: the image plane is divided into a $40{\times}25$ grid of cells; only the highest-confidence correspondence is considered in each cell, and the total number of correspondences per frame is capped at $M_\text{max}$.
\end{enumerate}
These operations improve robustness by removing invalid matches and by reducing spatial redundancy, so that the optimization is not dominated by densely clustered correspondences in only a small region of the image.
% Together, they discard unusable correspondences and limit spatial redundancy (per-cell selection and cap $M_\text{max}$), so no single image region dominates the fit.

\subsubsection{Confidence-Weighted Residuals}

Not all correspondences from Stage~1 should contribute equally to the final solution. To take their varying reliability into account, we scale each reprojection residual by a weight $w_{ik}$ computed from the Stage~1 confidence score $c_{ik}$:
\begin{equation}
    r_{ik} = w_{ik}\,\bigl(\bar{\mathbf{u}}_{ik} - \pi_i(T_i,\,\mathbf{p}_k)\bigr)
\end{equation}
where $w_{ik}$ is a monotonically increasing function of $c_{ik}$. This formulation is equivalent to generalized least squares with $\sigma^2_{ik} \propto 1/w_{ik}^2$, meaning that high-confidence matches receive greater influence while uncertain correspondences are naturally down-weighted.

\subsubsection{Two-Pass Solver}

To further improve robustness, we solve the optimization in two passes:
\begin{enumerate}
    \item \textbf{Pass~1}: minimize Eq.~\eqref{eq:objective} over all preprocessed correspondences until convergence.
    \item \textbf{Outlier gate}: remove correspondences whose unweighted reprojection error exceeds $\tau$ pixels.
    \item \textbf{Pass~2}: re-run the solver from the Pass-1 solution using only the retained inlier set, yielding the final optimized extrinsics $T_1^*,\, T_2^*$.
\end{enumerate}

This fit-gate-refit strategy allows the first pass to establish a geometrically coherent solution using reprojection together with the prior term, then uses that intermediate solution to identify and reject remaining gross outliers before the final refinement. Optimization is performed using the Trust-Region Reflective (TRF) Solver.

\section{Experimental Results}
\label{sec:experiments}

This study is organized around four experimental configurations, two datasets with different difficulty profiles, and metrics that capture both per-camera calibration accuracy and inter-camera consistency, i.e., how close the estimated camera-to-camera transformation is to the ground-truth camera-to-camera transformation.

\subsection{Experimental configurations}

We evaluate four configurations to isolate the effect of each stage:

\begin{itemize}
    \item \textbf{Stage-1 only (CMRNext)}: CMRNext predictions are used directly, without BA.
    \item \textbf{BA with reprojection only}: Stage-2 BA is applied using only the reprojection term ($\lambda = 0$, $\mu = 0$).
    \item \textbf{Relative and per-camera priors only (no reprojection)}: Stage-2 BA uses the $(\lambda,\mu)$ for the relative and per-camera priors, but the reprojection term is omitted from the objective.
    \item \textbf{Full method}: Stage-2 BA with all objective terms active: reprojection, per-camera prior, and relative-pose prior ($\lambda > 0$, $\mu > 0$).
\end{itemize}

To study the effect of temporal accumulation, we run each configuration on randomly sampled frame subsets of different lengths. We use $N \in \{10,\, 100\}$ on the KITTI and Walkley datasets.

\subsection{Initialization}

All experiments use a perturbed initial extrinsic: the ground-truth transform is displaced by
\begin{equation}
    T_\text{init} = T_\delta \cdot T_\text{GT}
\end{equation}
where $T_\delta$ applies a translation of magnitude $1.5\,\text{m}$ and a rotation of $20^\circ$, with components randomly sampled across all three axes.
This matches the large-misalignment regime commonly used in prior work \cite{schneider2017regnet, lv2021lccnet, cattaneo2019cmrnet} and tests whether the pipeline can recover accurate calibration from a realistically difficult starting point.

\subsection{Evaluation Metrics}

\paragraph{Extrinsic error:}
For each camera $C_i$, the estimated extrinsic $T_{L\to C_i}^\text{est}$ is compared against the reference transform $T_{L\to C_i}^\text{ref}$. Rotation and translation errors are computed as

\begin{align}
    \Delta R_i &= R_{L\to C_i}^\text{est}\,\bigl(R_{L\to C_i}^\text{ref}\bigr)^\top, \quad
    \varepsilon_R^i = \angle(\Delta R_i) \quad [\text{deg}], \\
    \varepsilon_t^i &= \bigl\|\mathbf{t}_{L\to C_i}^\text{est} - \mathbf{t}_{L\to C_i}^\text{ref}\bigr\|_2 \quad [\text{cm}].
\end{align}

\paragraph{Inter-camera consistency:}
The estimated inter-camera pose implied by the estimated extrinsics is
\begin{equation}
    T_{C_1 \to C_2}^\text{est} = T_{L\to C_2}^\text{est}\,\bigl(T_{L\to C_1}^\text{est}\bigr)^{-1}
\end{equation}
This is compared against the reference inter-camera pose $T_{C_1 \to C_2}^\text{ref}$ using the same rotation and translation error measures.

\paragraph{Reprojection error:}
For each correspondence, the reprojection residual is
\begin{equation}
    e_{ik} = \bigl\|\bar{\mathbf{u}}_{ik} - \pi_i\!\left(T_{L\to C_i}^\text{est},\,\mathbf{p}_k\right)\bigr\|_2 \quad [\text{px}]
\end{equation}
We report the median reprojection error and the improvement relative to Stage-1, $\Delta e_\text{med} = e_\text{med}^\text{Stage1} - e_\text{med}^\text{Stage2}$.

\subsection{Datasets}
In this work, we consider two datasets: the KITTI and Walkley datasets.
\subsubsection{KITTI}
The KITTI odometry benchmark \cite{geiger2013kitti} provides synchronized measurements from a Velodyne HDL-64E LiDAR and a calibrated stereo camera pair. We use Sequence~00, which contains 4541 frames of urban driving with buildings, roads, parked vehicles, and vegetation. Camera~2 (left color, $1242{\times}375$\,px) is treated as the primary camera and Camera~3 (right color, same resolution) as the secondary. Ground-truth extrinsics are provided with the sequence.

KITTI serves as the in-domain evaluation case. Since the scene content, sensing geometry, and calibration setting are close to those used by current learning-based methods, Stage-1 is already expected to be strong. In this dataset, the role of Stage-2 is to determine whether joint refinement can still reduce residual error and improve inter-camera consistency. 

\subsubsection{Walkley}
The Walkley dataset is a custom multi-sensor outdoor capture collected in Ottawa, Ontario, Canada \cite{OurData}. The rig consists of two forward-facing color cameras ($1920{\times}1200$\,px) and one Velodyne LiDAR. The sequence includes buildings, vegetation, parked vehicles, and powerlines, with vehicle motion covering straight segments, turns, stops, and speed changes. The full dataset contains 226 synchronized frames.

Walkley is used as the out-of-domain evaluation case. The sensor configuration, image resolution, and capture conditions differ from those seen by Stage-1, making it a useful test of whether the joint formulation can recover from weaker single-camera predictions.

\subsection{Ground Truth Calibration}

\paragraph{KITTI:}
Reference camera-LiDAR extrinsics are taken directly from the KITTI Sequence~00 calibration files.

\paragraph{Walkley:}
Since factory calibration was not available for the Walkley rig, reference extrinsics were established by manual point picking. For each camera, a synchronized image-point-cloud pair was loaded into an interactive dual-viewer. The operator selected a 3D LiDAR point in Open3D and then identified the corresponding pixel in the image. At least $N_\text{pp} \geq 6$ correspondences were collected per camera, distributed across spatially separated scene structures. The extrinsic was estimated using \texttt{cv2.solvePnPRansac}, testing multiple PnP solvers (Iterative, EPnP, SQPnP), and the solution with the lowest mean reprojection error was retained. The final calibration was accepted after visual overlay validation on utility poles, building edges, and vehicle boundaries.

\subsection{Implementation Details}

Stage-2 optimization is implemented in Python using \texttt{scipy.optimize.least\_squares} with the TRF method. A single static extrinsic per camera is estimated jointly over all $N$ frames; no per-frame poses are refined. Convergence tolerances are set to $10^{-6}$ for changes in objective value and parameter values, with a maximum of 2000 function evaluations.

\paragraph{Hyperparameters.}
Prior weights $(\lambda,\mu)$ and the robust loss scale $\delta$ were selected per dataset by random search over discrete candidate sets, minimizing a combined rotation-and-translation error objective evaluated on the full method. This yields $(\lambda,\mu) = (1,\,5)$ for KITTI and $(2,\,10)$ for Walkley, with Cauchy scale $\delta = 4$\,px and $8$\,px respectively. The outlier gate threshold is $\tau = 3$\,px for KITTI and $16$\,px for Walkley; the larger value on the out-of-domain dataset accommodates the wider residual spread observed from Stage-1 on Walkley.

\paragraph{Confidence weighting and correspondence cap.}
On KITTI, where per-frame confidence is high and uniform, confidence weights are uniform ($w_{ik} = 1$). On Walkley, we use $w_{ik} = \sqrt{\mathrm{clip}(c_{ik},\,0.1,\,1.0)}$, clipping the score to a minimum of 0.1 and a maximum of 1.0 before taking the square root. The hard minimum threshold is $c_\text{min} = 0.1$ (KITTI) and $0.2$ (Walkley). Correspondence subsampling retains at most one match per cell of a $40{\times}25$ image grid; the resulting effective cap of 1000 correspondences per frame is well below the $M_{\max} = 10{,}000$ hard limit.

All experiments were run on an Intel Core i7-12700H with an NVIDIA GeForce RTX 3060 Laptop GPU (32\,GB RAM). Stage-1 runs on GPU; Stage-2 runs on CPU and adds less than 5\% of total runtime in all configurations. Table~\ref{tab:runtime} summarises the breakdown.

\begin{table}[h]
\centering
\caption{Pipeline runtime breakdown. Stage-1 totals assume $0.9$\,s/frame (KITTI) and $3.2$\,s/frame (Walkley). Stage-2 timings are measured wall-clock for one pass of the joint BA.}
\label{tab:runtime}
\footnotesize
\setlength{\tabcolsep}{5pt}
\renewcommand{\arraystretch}{1.1}
\begin{tabular}{l c r r}
\toprule
\textbf{Stage} & \textbf{Dataset} & $N{=}10$ & $N{=}100$ \\
\midrule
\multirow{2}{*}{Stage-1} & KITTI   & ${\approx}9$\,s   & ${\approx}90$\,s  \\
                                & Walkley & ${\approx}32$\,s  & ${\approx}320$\,s \\
\midrule
\multirow{2}{*}{Stage-2} & KITTI   & ${\approx}0.5$\,s & ${\approx}5$\,s   \\
                                & Walkley & ${\approx}0.5$\,s & ${\approx}3$\,s   \\
\bottomrule
\end{tabular}
\end{table}

\subsection{Experiments Across Pipeline Configurations}

We first compare the four configurations to analyze how much of the final performance improvement is due to the reprojection-based refinement alone, how much is contributed by the relative and per-camera prior terms (reprojection-only BA to full method), and what happens when only the priors remain (no reprojection term).

\subsubsection{KITTI}

Table~\ref{tab:kitti_metrics} reports inter-camera consistency and median reprojection error on KITTI. With $N{=}10$, Stage-1 already produces low reprojection error and near-zero inter-camera rotation error. In this case, Stage-2 provides only limited room for improvement, which is expected given that the initialization is already close to the reference solution. Even so, configurations that include reprojection reduce median reprojection error, and the full method gives the lowest inter-camera translation error.

At $N{=}100$, the effect of joint refinement becomes more obvious. Stage-1 only yields an inter-camera rotation error of 0.206$^\circ$; reprojection-based BA improves this metric substantially, and the full method gives the best rotational consistency and the lowest reprojection error on both cameras. The gain of the full method over reprojection-only BA is modest on KITTI, but consistent with the operating regime: when Stage-1 is already strong, the prior terms mainly act as stabilizers rather than as recovery mechanisms.

\begin{table}[t]
\centering
\caption{Inter-camera consistency and median reprojection error on KITTI Sequence~00. IC-$\Delta R$/$\Delta t$: inter-camera rotation/translation error.}
\label{tab:kitti_metrics}
\footnotesize
\setlength{\tabcolsep}{4pt}
\renewcommand{\arraystretch}{1.1}
\resizebox{\columnwidth}{!}{%
\begin{tabular}{l c c c c c}
\toprule
\textbf{Method} & $N$ & IC-$\Delta R$ [$^\circ$] & IC-$\Delta t$ [cm] & \makecell{Reproj \\ Primary [px]} & \makecell{Reproj \\ Secondary [px]} \\
\midrule
CMRNext (Stage-1) & 10  & \textbf{0.001} & 5.04 & 1.09 & 0.52 \\
BA (reproj.\ only) & 10  & 0.035 & 4.37 & 0.72 & 0.43 \\
\makecell[l]{BA (rel.-pose + per-cam. only)} & 10  & 0.178 & 4.18 & 2.19 & 0.56 \\
Full method & 10  & 0.033 & \textbf{4.11} & \textbf{0.67} & \textbf{0.41} \\
\midrule
CMRNext (Stage-1) & 100 & 0.206 & 3.56 & 2.57 & 0.65 \\
BA (reproj.\ only) & 100 & 0.040 & 4.46 & 0.77 & 0.54 \\
\makecell[l]{BA (rel.-pose + per-cam. only)} & 100 & 0.178 & 4.18 & 2.19 & 0.56 \\
Full method & 100 & \textbf{0.037} & \textbf{4.16} & \textbf{0.71} & \textbf{0.50} \\
\bottomrule
\end{tabular}
}
\end{table}

\subsubsection{Walkley}

Table~\ref{tab:walkley} shows the same progression on Walkley, where the Stage-1 predictions are much less reliable. Here the differences between configurations are larger and easier to interpret.

Stage-1 produces large translation errors on both cameras, reaching 87.2\,cm (primary) and 108.6\,cm (secondary) at $N{=}100$. The main recovery step is reprojection-only BA, which reduces secondary translation error to 3.57\,cm and primary to 25.7\,cm, demonstrating that multi-frame reprojection fitting is the primary recovery mechanism on out-of-domain data. Moving to the full method provides further incremental improvement across all metrics. At $N{=}100$, primary translation error drops to 22.6\,cm, secondary to 3.14\,cm, and inter-camera translation error decreases from 23.9\,cm to 21.0\,cm.

\textbf{BA (relative- and per-camera priors only, no reprojection).} This configuration remains far from both reprojection-only BA and the full method: at $N{=}100$, primary translation error is 75.6\,cm and secondary 93.2\,cm, i.e., only slightly better than Stage-1 only and worse than reprojection-only BA. This confirms that the large gains come from fitting the learned 2D--3D correspondences, not from the prior terms alone. The full method therefore remains the strongest configuration.

This pattern is consistent across frame counts. The main practical effect of the full objective is lower inter-camera inconsistency and better absolute calibration, especially for the weaker view. On Walkley, that is the secondary camera, whose Stage-1 error is the largest and whose improvement under the full method is the most pronounced. This is the system where the relative-pose prior is most useful: the optimizer can use information from the better-constrained view without letting either camera drift arbitrarily far from the Stage-1 solution.

\begin{table*}[t]
\centering
\caption{Calibration results on the Walkley dataset. Priors only (no reprojection) uses $(\lambda,\mu)$ as in the full method but omits the reprojection term (control without data fitting). Bold: best per column block.}
\label{tab:walkley}
\footnotesize
\setlength{\tabcolsep}{2.5pt}
\renewcommand{\arraystretch}{1.1}
\begin{tabular}{l c | c c c c c c c c | c c c c c c c c | c c}
\toprule
& & \multicolumn{8}{c|}{\textbf{Primary Camera}} & \multicolumn{8}{c|}{\textbf{Secondary Camera}} & \multicolumn{2}{c}{\textbf{Inter-Camera}} \\
\cmidrule(lr){3-10}\cmidrule(lr){11-18}\cmidrule(lr){19-20}
\textbf{Method} & $N$ & $E_t$ & $E_R$ & $x$ & $y$ & $z$ & roll & pitch & yaw & $E_t$ & $E_R$ & $x$ & $y$ & $z$ & roll & pitch & yaw & IC-$\Delta R$ & IC-$\Delta t$ \\
\midrule
CMRNext (Stage-1) & 10  & 150.9 & 0.91  & 51.0 & 0.97 & 142.0 & 0.21 & 0.51 & 0.74 & 41.6  & 0.63  & 24.9 & 22.2 & 24.8  & 0.32 & 0.28 & 0.49 & 1.08  & 117.1 \\
BA (reproj.\ only) & 10  & 60.2  & 1.06  & 10.5 & 30.7 & 50.7  & 0.46 & 0.39 & 0.87 & 24.3  & 0.50  & 1.23 & 24.0 & 3.76  & 0.35 & 0.06 & 0.35 & 1.06  & 47.8  \\
\makecell[l]{BA (rel.-pose + per-cam. only)} & 10 & 133.7 & 0.80 & 45.7 & 0.83 & 125.6 & 0.18 & 0.44 & 0.64 & 35.5 & 0.56 & 21.4 & 19.9 & 21.3 & 0.28 & 0.25 & 0.43 & 1.06 & 100.9 \\
Full method & 10  & 54.2  & 0.95  & 9.50 & 27.6 & 45.7  & 0.41 & 0.35 & 0.78 & 21.9  & 0.45  & 1.10 & 21.6 & 3.37  & 0.32 & 0.06 & 0.31 & 0.95  & 43.1  \\
\midrule
CMRNext (Stage-1) & 100 & 87.2  & 1.09  & 24.8 & \textbf{13.2} & 82.5  & 0.35 & 0.48 & 0.89 & 108.6 & 1.66  & 2.46 & 27.9 & 104.9 & 0.64 & 0.09 & 1.54 & 1.06  & 87.0  \\
BA (reproj.\ only) & 100 & 25.7  & 0.825 & 13.8 & 16.8 & 13.8  & 0.27 & 0.32 & 0.71 & 3.57  & 0.230 & 2.31 & 0.31 & 2.70  & 0.07 & 0.05 & 0.21 & 0.914 & 23.9  \\
\makecell[l]{BA (rel.-pose + per-cam. only)} & 100 & 75.6 & 0.9 & 22.2 & 15.3 & 71.7 & 0.31 & 0.41 & 0.79 & 93.2 & 1.5 & 2.2 & 23.9 & 93.7 & 0.57 & 0.08 & 1.3 & 0.9 & 76.5 \\
Full method & 100 & \textbf{22.6} & \textbf{0.726} & \textbf{12.1} & 14.8 & \textbf{12.1} & \textbf{0.24} & \textbf{0.29} & \textbf{0.62} & \textbf{3.14} & \textbf{0.203} & \textbf{2.03} & \textbf{0.27} & \textbf{2.38} & \textbf{0.06} & \textbf{0.04} & \textbf{0.19} & \textbf{0.804} & \textbf{21.0} \\
\bottomrule
\end{tabular}
\end{table*}

\subsection{Comparison with Prior Methods on KITTI}

After isolating the effect of each pipeline stage, we compare the final system against published camera--LiDAR calibration methods on KITTI. Table~\ref{tab:kitti_comparison} reports results under the standard $\pm$1.5\,m\,/\,$\pm$20$^\circ$ perturbation. We include both the Stage-1 baseline and the full method for $N{=}10$ and $N{=}100$ frames on Sequence~00.

At $N{=}10$, the Stage-1 baseline is already competitive. The full method provides only limited additional gain in this low-data, in-domain case. At $N{=}100$, the advantage of the full pipeline is more visible. The primary camera reaches 0.89\,cm translation error and 0.038$^\circ$ rotation error, improving over Stage-1 (2.63\,cm, 0.187$^\circ$) by roughly $3\times$ in translation and $5\times$ in rotation. The secondary camera also improves, and the full method outperforms all listed prior approaches on both cameras.

These results indicate that the proposed formulation does not trade rig consistency for per-camera accuracy. On KITTI, where learned single-camera matching is already effective, the joint refinement still improves the final calibration when enough frames are available.

\begin{table*}[t]
\centering
\caption{Calibration error on Sequence~00 of the KITTI dataset. $E_t$: translation magnitude [cm]; $E_R$: rotation magnitude [$^\circ$]; $x$/$y$/$z$: per-axis translation errors [cm]; roll/pitch/yaw: per-axis rotation errors [$^\circ$] (absolute values). Best results are in \textbf{bold}; second-best results are \underline{underlined}.}
\label{tab:kitti_comparison}
\footnotesize
\setlength{\tabcolsep}{2.5pt}
\renewcommand{\arraystretch}{1.1}
\begin{tabular}{l c | c c c c c c c c | c c c c c c c c}
\toprule
& & \multicolumn{8}{c|}{\textbf{Left Camera (Primary)}} & \multicolumn{8}{c}{\textbf{Right Camera (Secondary)}} \\
\cmidrule(lr){3-10}\cmidrule(lr){11-18}
\textbf{Method} & \textbf{Init} & $E_t$ & $E_R$ & $x$ & $y$ & $z$ & roll & pitch & yaw & $E_t$ & $E_R$ & $x$ & $y$ & $z$ & roll & pitch & yaw \\
\midrule
LCCNet~\cite{lv2021lccnet}   & $\pm$1.5m/$\pm$20$^\circ$ & \underline{1.01}  & 0.12  & \textbf{0.26} & \underline{0.36} & 0.35 & \underline{0.02} & 0.11 & \textbf{0.03} & 52.51 & 1.47  & 52.48 & \textbf{0.26}  & \underline{0.74} & \textbf{0.01} & 1.47 & \underline{0.03} \\
RGGNet~\cite{yuan2020rggnet}   & $\pm$0.3m/$\pm$20$^\circ$ & 11.49 & 1.29  & 8.14 & 2.79 & 3.97 & 0.35 & 0.74 & 0.64 & 23.52 & 3.87  & 18.03 & 5.55  & 6.06 & 0.51 & 3.38 & 1.48 \\
CMRNet~\cite{cattaneo2019cmrnet}   & $\pm$1.5m/$\pm$20$^\circ$ & 1.57  & 0.10  & 1.06 & 0.74 & 0.34 & 0.03 & \textbf{0.01} & 0.08 & 52.92 & 1.49  & \underline{1.59}  & 52.87 & \textbf{0.36} & 0.04 & \underline{0.02} & 1.49 \\
CMRNext~\cite{cattaneo2025cmrnext}  & $\pm$1.5m/$\pm$20$^\circ$ & 1.89  & 0.08  & 1.12 & 0.83 & 0.79 & 0.04 & 0.04 & \underline{0.04} & 7.07  & 0.23  & \textbf{2.17}  & 5.78  & 0.94 & 0.05 & 0.05 & 0.20 \\
\midrule
Stage-1 ($N$=10)${}$  & $\pm$1.5m/$\pm$20$^\circ$ & 1.14 & \textbf{0.002} & 1.12 & \textbf{0.07} & \underline{0.22} & \textbf{0.01} & 0.04 & 0.10 & 5.99 & \textbf{0.001} & 5.84 & 0.97 & 0.95 & 0.04 & \underline{0.02} & \textbf{0.01} \\
\textbf{Ours} ($N$=10)${}$  & $\pm$1.5m/$\pm$20$^\circ$ & 1.19 & 0.043 & 1.06 & 0.53 & \textbf{0.11} & \underline{0.02} & \underline{0.02} & \textbf{0.03} & \underline{5.17} & \underline{0.030} & 5.12 & \underline{0.57} & \underline{0.51} & 0.03 & \textbf{0.00} & \textbf{0.01} \\
Stage-1 ($N$=100)${}$ & $\pm$1.5m/$\pm$20$^\circ$ & 2.63 & 0.187 & 2.58 & 0.46 & 0.21 & 0.04 & 0.09 & 0.08 & 6.23 & 0.137 & 6.13 & 0.78 & 0.76 & 0.04 & 0.03 & \underline{0.02} \\
\textbf{Ours} ($N$=100)${}$ & $\pm$1.5m/$\pm$20$^\circ$ & \textbf{0.89} & \underline{0.038} & \underline{0.77} & \underline{0.34} & \underline{0.20} & \textbf{0.01} & \textbf{0.01} & \textbf{0.03} & \textbf{4.97} & \underline{0.030} & \underline{4.88} & 0.67 & 0.71 & \underline{0.01} & \underline{0.01} & \textbf{0.01} \\
\bottomrule
\end{tabular}
\end{table*}

\subsection{Qualitative Results}

\begin{figure*}[t]
    \centering
    \includegraphics[width=0.9\linewidth]{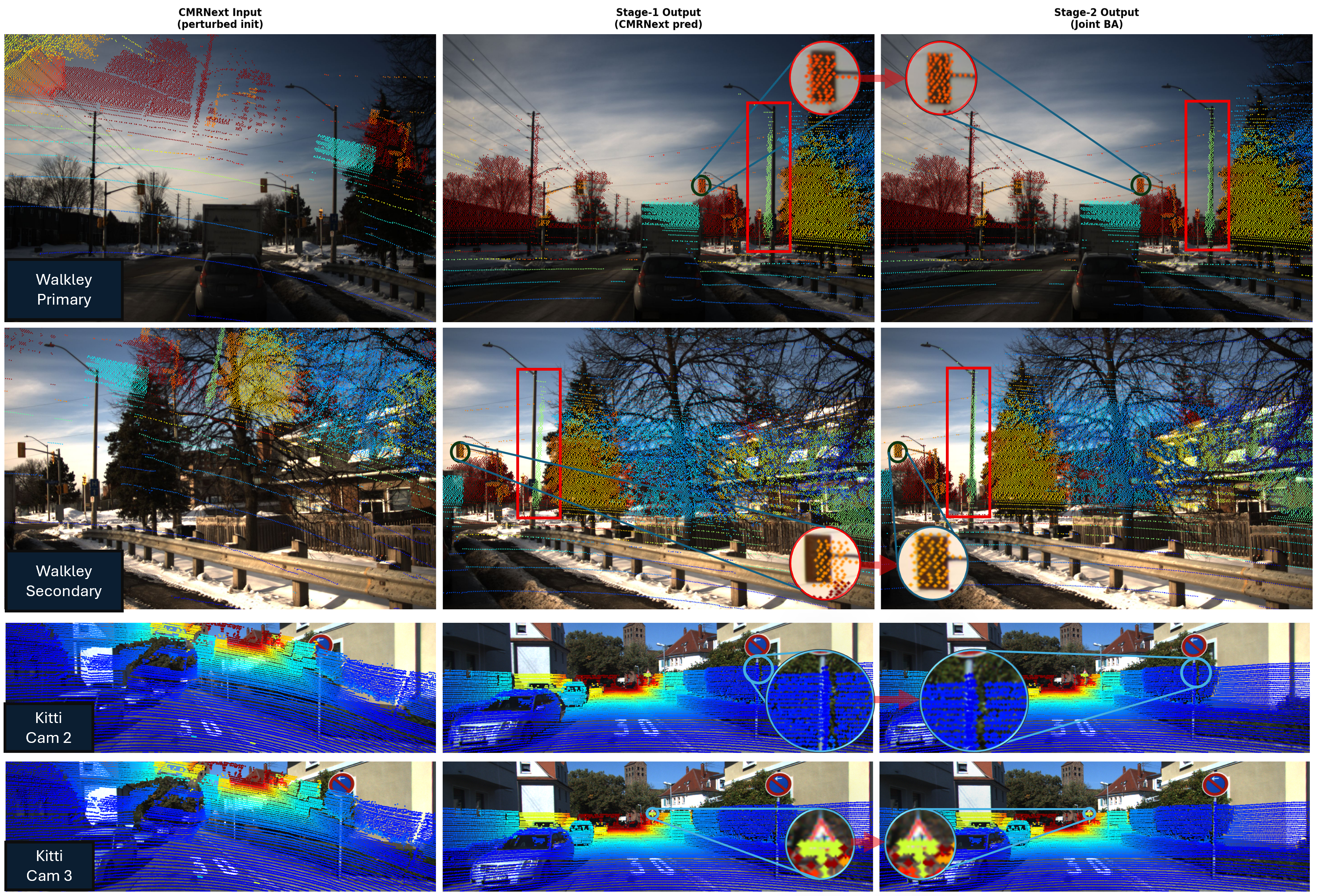}
    \caption{LiDAR projection overlays from Walkley and KITTI. Columns show the perturbed initialization used as input to CMRNext, the Stage-1 output, and the Stage-2 output after joint BA for the $N{=}100$ frames}
    \label{fig:qualitative}
\end{figure*}

Figure~\ref{fig:qualitative} shows representative projection overlays from both Walkley and KITTI. The qualitative behavior is consistent with the quantitative results and differs across the two datasets.

On Walkley, the effect of Stage-2 is visually clear in both cameras. Starting from the perturbed initialization, Stage-1 recovers a plausible alignment, but residual offsets remain on prominent vertical structures. After joint BA, these offsets are reduced across the scene. The highlighted traffic lights and light pole show the clearest improvement. This stronger visual gain is consistent with the larger numerical improvement on Walkley, where the Stage-1 predictions are weaker and the joint optimization has more room to correct them.

On KITTI, the visual difference between Stage-1 and Stage-2 is more subtle. This is expected, since the Stage-1 predictions are already accurate on this in-domain dataset. For Camera~2, the improvement is visible on the highlighted roadside pole, where the Stage-2 overlay follows the pole boundary more closely. For Camera~3, the refinement is harder to observe and becomes clearer only on close inspection of smaller, distant structures. In the highlighted example, the improvement appears on the sign farther down the road, approximately 40\,m from the camera, where Stage-2 better aligns the projected points with the sign edges. The full method still improves the solution, but the gain is smaller and often appears only on fine structures or at longer range.

\section{Conclusion}

We presented a two-stage framework for joint multi-camera LiDAR extrinsic calibration that refines learned single-camera predictions through a multi-frame bundle adjustment with inter-camera consistency priors.

On KITTI, the full method improves the final calibration and achieves 0.89\,cm translation error and 0.038$^\circ$ rotation error on the primary camera at $N{=}100$. The relative-pose and per-camera priors also improve inter-camera consistency over reprojection-only refinement. On Walkley, multi-frame reprojection BA reduces the secondary-camera translation error from 108.6\,cm to 3.57\,cm; the relative-pose prior further refines this to 3.1\,cm while also improving inter-camera consistency and absolute accuracy on the primary camera.

Several limitations remain. When few frames are available and CMRNext is already near its best performance on in-domain data, the proposed BA method offers limited improvement, as observed on KITTI at $N{=}10$. On Walkley, residual primary camera translation errors (${\sim}22$\,cm at $N{=}100$) indicate that out-of-domain correspondences remain a limiting factor that the optimization cannot fully recover from. At the same time, the improvements from 87.2 cm translation error to 22 cm show that cross-camera information through BA is beneficial when the initial pairwise estimates are less reliable. The priors-only configuration corroborates that these gains require the reprojection data term: the prior terms alone do not recover large Stage-1 errors. Future work includes extending the framework to rigs with more than two cameras, incorporating an online receding-window bundle adjustment for continuous re-calibration, and exploring end-to-end training of the correspondence network jointly with the geometric refinement objective.

\bibliographystyle{IEEEtran}
\bibliography{references}

\end{document}